# Preparation of Improved Turkish DataSet for Sentiment Analysis in Social Media

*Semiha* Makinist[1,*], *İbrahim Rıza* Hallaç[2,*], *Betül* Ay Karakuş[2,*], and *Galip* Aydın[2,*]

[1] Sentis Software  
[2] Department of Computer Engineering, Firat University, Elazig, Turkey

**Abstract.** A public dataset, with a variety of properties suitable for sentiment analysis [1], event prediction, trend detection and other text mining applications, is needed in order to be able to successfully perform analysis studies. The vast majority of data on social media is text-based and it is not possible to directly apply machine learning processes into these raw data, since several different processes are required to prepare the data before the implementation of the algorithms. For example, different misspellings of same word enlarge the word vector space unnecessarily, thereby it leads to reduce the success of the algorithm and increase the computational power requirement. This paper presents an improved Turkish dataset with an effective spelling correction algorithm based on Hadoop [2]. The collected data is recorded on the Hadoop Distributed File System and the text based data is processed by MapReduce programming model. This method is suitable for the storage and processing of large sized text based social media data. In this study, movie reviews have been automatically recorded with Apache ManifoldCF (MCF) [3] and data clusters have been created. Various methods compared such as Levenshtein and Fuzzy String Matching have been proposed to create a public dataset from collected data. Experimental results show that the proposed algorithm, which can be used as an open source dataset in sentiment analysis studies, have been performed successfully to the detection and correction of spelling errors.

## 1 Introduction

Today, most of the social media sources we meet frequently encounter in a very wide range of fields are generate by video sharing (YouTube), photo sharing (Instagram), and location based applications (Foursquare), blogs, microblogs (Twitter), social networks (Facebook)... Social media, made up of these different digital media, has become an important source of information to many researchers in the form of constantly updated information on many issues such as popular events, politics, products, or services. For example, a politician can use this data to make a conclusion about what people think about him/her when conducting an election campaign. Similarly, a company may collect data from social media for analyzing the opinions of their customers on their products.

According to March 2015 data, Facebook has more than 1 billion registered users, and Twitter has 288 million active users per month [4]. Linkedin, Tumblr, Instagram and many other social networks are increasing their number of users every day. The need to store, analyze and visualize data in social networks is not only a problem of these social networking companies. In many different areas, people and organizations want to access social media data and analyze it from various perspectives.

The collection, management and analysis of these data is a major big data problem [5]. Many studies have been done in this field in the literature and usually Twitter is preferred as the social media platform. One of the most important reasons for this is that Twitter users share short and many messages. By benefitting from these social media posts one can come up with the distribution of positive and negative thoughts, the tendencies of the targeted customer groups, the reputation and influence status on social media for people or companies. These type of analyses are mainly carried out by applying machine learning techniques on large volumes of data. Different methods are used during the collection of data such as data can be collected by using the application programming interface (Twitter API, Facebook API, etc.) of the platform, can be saved manually, or a problem specific data collection method can be used.

---

[*] Semiha Makinist: semihamakinist@gmail.com  
[*] İbrahim Rıza Hallaç: irhallac@firat.edu.tr  
[*] Beetül Ay Karakuş: betulay@firat.edu.tr  
[*] Galip Aydın: gaydin@firat.edu.tr

In the literature, Pak and Paroubek [6] created a data structure for automatic sentiment analysis to predict if a newly added data is positive, negative, or neutral. Albert and Eibe [7] proposed a sliding window Kappa statistic for evaluation in time-changing data stream and they dynamically classify the instant data flows on Twitter. Yerve et al. [8] detects in microblogs whether there were posts related to difficult companies, such as "Apple, BlackBerry". Roberto, Smaranda and Nina [9] analyzed tweets in their study to analyze existence of sarcasm. Agarwal et al. [10] performed sentiment analysis of tweets using a tree kernel method. Saif et al. [11] obtained quite good results by adding additional semantic features (eg, the feature of the device on which the tweet is inserted) to the classification methods to identify the sentiment of the tweets that are posted by the companies or the individuals. Spencer and Uchyigit [12] used a web-based tool which is called as "Sentimator" to collect data from Twitter in real-time according to specified search terms, and they developed a sentiment analysis tool using the Naive Bayes classification algorithm and natural language processing libraries. Kouloumpis, Wilson and Moore [13] performed a classification task by using the hashtags and emoticons written in the tweets along with the consideration of the linguistic features a language. Saif and his friends [14] formed two different clusters in their studies to reduce the density of the Twitter data which is to be used for sentiment analysis. Agarwal and Sabharwal [15] proposed a sentiment analysis technique for Twitter texts which uses an end-to-end pipeline method to group tweets into four groups; positive, negative, objective and neutral categories. Cody et al. [16] tried to identify climatic changes by analyzing data collected from Twitter. Paltoglou [17] conducted a study on emotion-based event detection. Singhal and his friends [18] used Twitter data to predict which party would win the elections.

In the mentioned studies, sentiment analysis and other prediction based applications were performed using textual data. The quality of the data sets used in these studies is highly critical and it affects the success of the results at high rates. In addition, the size of the word-vector-matrices, which are input data of the text-based classification and clustering algorithms such as TF-IDF, are greatly increased as a result of deriving different words because of the wrong spellings of the same word. This increase, significantly reduces the success of the machine learning algorithm. In addition, every word added will increase the required processing capacity and reduce the computational performance. In this study we propose a method for creating a data set in Turkish to improve the performance of sentiment analysis studies which are based on textual data. Various improvement studies have been carried out in order to increase the data quality and different methods have been tried. Some problems have been encountered in the experiments and some solutions have been proposed for solving these problems. In the first part of the article, the introduction is given, in the second part how the data is gathered, in the third part how the data set is formed, in the fourth section the proposed algorithm is explained. In the fifth part we compare our method with other methods in the literature. In the sixth part we share the experimental results.

## 2 Data Collection

Obtaining the data covers a significant part of this work. User comments in a movie site are preferred as data source. There are also ratings on the movie site, with comments from users and ratings of 1 and 5 points. The reason for choosing film critics as datasets is that there are numerical rating values given along with the user comments.

Apache ManifoldCF (MCF) was used to obtain the data. MCF technology is an application that transfers data contained in data sources to target resources. In the general use of MCF, data mapping is done by establishing a bridge to content indexing systems from enterprise content management systems (also applying content security policies during the process).

In this study, MCF was used as a web browser and user comments in the designated movie site were drawn to the file system. The first step in collecting data with MCF is to set the source from which the data will be pulled. In MCF, this is called repository connector and there are many repository connectors in MCF. One of these is the web connector. The web connector passes the data of the web pages to the target sources, and if any authorization system is used before this connector is created, an appropriate idle authorization connector must be created. Using this technology, the data is drawn with the following sequence:

1. Create an empty authorization group.
2. Create a Web adapter.
3. A time adapter is created to configure which time intervals the data will be retrieved.
4. Create a file adapter to save the data.
5. After all the necessary adapters have been created, finally a transfer job needs to be created. In other words, which site to connect to and the index area (user comments) is determined.
6. Once all the work is done, the created job is started.

Once all the configurations are complete, Apache ManifoldCF scans all web pages and saves the pages with user comments to the local file system in HTML format. In this way, 2983 movie reviews were collected for use as sample data sets. In order to extract the comments out of this data, the Groovy script language which can work on the JVM was used and the edited data was saved in JSON format.

## 3 Data Set Creation

The step of generating words or groups of words is a very important step in text-based analysis where natural language processing methods are used. While there are many resources and software libraries for creating these data sets in the literature, most of them are for the English language. There are very few methods and software libraries developed for Turkish. In this study, language processing and word base generation techniques have been proposed based on Turkish language [19].

The primary aim of the work done is to clear the comments correctly from unnecessary words and to create correct word lists from 1, 2 and 3 grams according to the scores given by the users. Another goal is to identify strong negative and strong positive word and word groups in word lists.

For this purpose, comments are first cleared of unnecessary words (conjunctions, prepositions, adverbs, special names, place names and country names), punctuation marks, and numbers. In the next step, we use the Zemberek library to check whether each word is in Turkish (language detection). If a word is misspelled in Turkish, the proposed method corrects the word and the word is converted to the correct one. As a result of these operations, 3 different files are obtained for each rating score between 0 and 5 points. A total of 15 files will be created in this way. From all these operations, lists of words in different grams in different types of ratings are obtained and they are saved in a file that is not in any other file, ie only in the file it is in. The main purpose of this step is to identify the most powerful words and phrases, especially those for the rating values of 0 and 5. Some of the results from the comments of 0 and 5 points of ratings for 1 gram is shown in Fig.1. Results are obtained by checking the spelling of the words with Zemberek and without making any additional change in the text. As you it can be seen in the figure, using the text without editing process is performed is not generating a very effective word list.

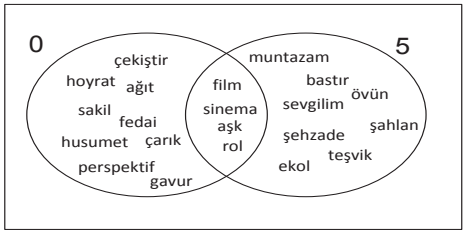

**Fig. 1.** 1-gram Turkish words from movie reviews with a star rating on 0-5 scale.

Some of the results that have been come into prominence on the list of words used and not used by Zemberek in 1 gram, which are obtained after correcting the movie reviews of 0 and 5 stars are shown in Fig. 2.

| 0 | 0&5 | 5 |
|---|---|---|
| iğrenç<br>nefret<br>gereksiz<br>izlemeyin<br>rezalet<br>sıkıcı<br>etmiyorum<br>berbat | film<br>sinemada<br>iyi<br>çok<br>kötü<br>kesinlikle<br>güzel<br>tavsiye | muhteşem<br>mükemmel<br>harika<br>etkileyici<br>süper<br>efsane<br>gerçekten<br>ederim |

(a)

| 0 | 0&5 | 5 |
|---|---|---|
| iğrenç<br>nefret<br>gerek<br>rezalet<br>sıkıcı<br>berbat | film<br>sinema<br>iyi<br>çok<br>kötü<br>kesin<br>güzel<br>tavsiye | muhteşem<br>harika<br>etki<br>süper<br>efsane<br>gerçek |

(b)

**Fig. 2.** The results from the comments of 0 and 5 stars of ratings for 1 gram (a) The results from the comments of 0 and 5 stars of ratings for 1 gram with Zemberek (b).

Some of the results that have been come into prominence on the list of words used and not used by Zemberek in 2 grams (Fig. 3) and 3 grams (Fig. 4).

| 0 | 0&5 | 5 |
|---|---|---|
| en kötü<br>berbat film<br>berbat ötesi<br>saçma sapan<br>tavsiye etmiyorum<br>vasat film<br>en berbat<br>yazıklar olsun<br>vakit kaybı<br>hak etmiyor | kötü film<br>çok kötü<br>son zamanlarda<br>tek kelimeyle<br>filmin sonu<br>kesinlikle<br>tavsiye | çok güzeldi<br>film iyi<br>görsel şölen<br>mutlaka izleyin<br>muhteşem film<br>süper film<br>baş yapıt<br>hak ediyor<br>çok beğendim<br>gerçekten harika<br>en iyi<br>çok kaliteli |

(a)

| 0 | 0&5 | 5 |
|---|---|---|
| en kötü<br>berbat film<br>berbat öte<br>saçma sapan<br>vasat film<br>en berbat<br>yazık ol<br>vakit kap | kötü film<br>çok kötü<br>son zaman<br>tek kelime<br>film son<br>kesin tavsiye<br>hak et<br>tavsiye et | çok güzel<br>film iyi<br>görsel şölen<br>mutlaka izle<br>muhteşem film<br>süper film<br>baş yapıt<br>çok beğen<br>gerçek harika |

(b)

**Fig. 3.** The results from the comments of 0 and 5 stars of ratings for 2 grams (a) The results from the comments of 0 and 5 stars of ratings for 2 grams with Zemberek (b).

| 0 | 0&5 | 5 |
|---|---|---|
| izlediğim en kötü<br>en berbat film<br>en kötü film<br>puanı hak etmiyor<br>verdiğim paraya<br>acırım<br>berbat ötesi film<br>gördüğüm en berbat<br>izlediğim en berbat<br>saçma sapan film<br>hayatımda böyle<br>saçma<br>rezalet ötesi film | hayatımda<br>izlediğim<br>en | izlediğim en iyi<br>en iyi film<br>kesinlikle izlenmesi<br>gereken<br>çok iyi film<br>şiddetle tavsiye ederim<br>herkese tavsiye ederim<br>mutlaka izlenmesi<br>gereken<br>gerçekten çok güzel<br>tek kelimeyle muhteşem<br>tam baş yapıt |

(a)

| 0 | 0&5 | 5 |
|---|---|---|
| izle en kötü<br>en berbat film<br>en kötü film<br>saçma sapan film<br>ver para acı<br>berbat öte film<br>gör en berbat<br>tek kelime berbat<br>izle en berbat<br>hayat böyle<br>saçma<br>rezalet öte film | hayat izle en<br>puan hak et<br>herkes tavsiye et | izle en iyi<br>en iyi film<br>kesin izle gerek<br>çok iyi film<br>şiddet tavsiye et<br>mutlaka izle gerek<br>gerçek çok güzel<br>tek kelime<br>muhteşem<br>tam baş yapıt |

(b)

**Fig. 4.** The results from the comments of 0 and 5 stars of ratings for 3 grams (a) The results from the comments of 0 and 5 stars of ratings for 3 grams with Zemberek (b).

It was found that the best results obtained from the studies were 2 and 3 grams word lists and it was observed that the word lists created by the proposed method compared to the word list created by using only Zemberek are more successful.

## 4 Proposed Method

In this study, some of the algorithms that have emerged in the literature such as Levenshtein Distance and Fuzzy String Matching have been tested. However, a new study was conducted because these methods did not achieve the desired success. The main goal of this study is to make:

- The words which are not written correctly in Turkish such as «cok, saglam, simarik, guzel» and
- Words with repetitive letters like «şoook, eeeen»,
- Abbreviations used in social media such as «kib, grrrz, slm, mrb, aeo, tmm» and
- To identify and correct spelling mistakes such as «gelrm, biliyrm, sonclr»

All of these steps are given below, respectively:

1. In the first step, it is checked whether there are any unnecessary repetitive characters in the word, if any, they are cleared. For example, a word "çoookkkk" is corrected to "çok".
2. In the next step, words with misspelled letters, such as "cok", "saglam" words, are converted into the correct Turkish writing in the form of "çok" and "sağlam". Different algorithms have been tried for this transformation process. For this purpose, *Levenshtein Distance*, *Fuzzy String Matching* and proposed word correction algorithm are used. The best results are obtained with the newly developed word correction algorithm.
3. Fuzzy String Matching Algorithm is used to detect letter errors. The test results are shown in Table 1 (the threshold value for all data is 0.8).
4. The Levenshtein Distance Algorithm is used to determine the word to be corrected according to the obtained results.

**Table 1.** Correcting letter errors with Fuzzy String Matching Algorithm.

| Word to fix | Correct Word | Results | |
|---|---|---|---|
| | | Output | Prob. |
| gelrm | gelirim | geldim | 0.83 |
| | | gelirim | 0.92 |
| | | geliyorum | 0.8 |
| | | germ | 0.8 |
| biliyrm | biliyorum | bildirim | 0.8 |
| | | bilim | 0.83 |
| | | biliyorum | 0.88 |
| sonuşlar | sonuçlar | sonuçlar | 0.88 |

In Table 1, a word list was used for the purpose of checking and this list was taken from "mythes-tr \ data" addressed by Anıl Özbek's GitHub account [20].

### 4.1 Word Correction Algorithm

A simple probability equation is used when the word correction algorithm is developed. With this algorithm, new words are derived by determining the letters that are likely to be corrected. For example, "cok, cök, çok, çök" words were expected to be derived from "çok" word, and it was confirmed that these words were derived as a result of the studies done. This process can be performed for more than one letter, and successful results up to 6 letters have been obtained. The formula for this algorithm is given in Eq. (1).

$$power = 2^n \qquad (1)$$

Table 2 describes the variables used in the equation used in the algorithm.

**Table 2.** Variable Definitions.

| Variables | Description |
|---|---|
| n | Number of elements to fix |
| 2 | Letter change status |
| power | Derived word count |

The operating principle of the proposed algorithm is shown in Fig. 5.

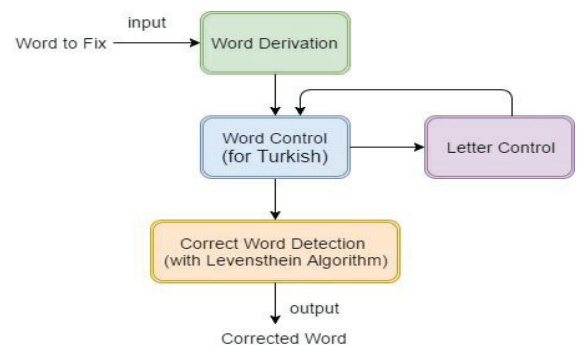

**Fig. 5.** The working principle of the word correction algorithm.

For example, the number of letters "2" to be changed in "çok" word (that is n = 2), the probability of each word is 2 since each word has two states, and the number of letters to be derived is calculated as 4 using the equation in (1).

As a result of the tests performed, it was observed that over 90% of the corrected words were achieved using the developed algorithm. The only significant disadvantage of this method is that the system slows down as the number of letters to be corrected increases. For example, if there are 18 letters in a word, this process takes approximately 1 hour. The reason for this is that (that is 262144) derivations of the word are done.

## 5 Experimental Results

The experimental results of the methods used to correct the words are given comparatively in the table context in this section. Several similarity algorithms have been used in the literature for this process, but in accordance with this study, tests have been performed using Levenshtein, Fuzzy String Matching and the proposed algorithm. The comparison of these algorithms as a result of the tests performed is given in Table 3. The threshold value for all tests performed in Table 3 was set at 0.6. In the Levenshtein Algorithm, it is assumed that the cost is closer to 0, so the similarity is closer to 1 in the Fuzzy String Matching Algorithm. A threshold value is not used for the word correction algorithm.

Table 3. The Experimental Results of the Methods.

| Algorithms | Test Data | Results | |
|---|---|---|---|
| | | Output | Prob. |
| Levenshtein | cok | cık | 0.33 |
| | | cop | 0.33 |
| | | cuk | 0.33 |
| | | çok | 0.33 |
| Fuzzy String Matching | cok | cık | 0.66 |
| | | cokey | 0.75 |
| | | coşarak | 0.66 |
| | | coşku | 0.75 |
| | | coşkulu | 0.6 |
| | | coşkun | 0.66 |
| | | coşmak | 0.66 |
| | | cuk | 0.66 |
| | | çok | 0.66 |
| Proposed Method | cok | çok | 1 |

As a result of observations made according to the test results, It has been found that using only Levenshtein and Fuzzy String Matching Algorithms is not sufficient. It is observed that these algorithms are better used in the correct place in the word correction algorithm.

## 6 Conclusion

Although data warehouses and open-source libraries for many languages are included in the literature, a full Turkish data set and library are not open source. Therefore, open source data sets and libraries are needed in Turkish sentiment analysis and other natural language processing studies. Moreover, the storage of social media data and the high cost of calculation make it a major problem. These problems can be eliminated through frameworks that support distributed programming such as Hadoop, Spark. For this purpose, the paper presented an improved Turkish dataset with an effective spelling correction algorithm with Hadoop Distributed File System and MapReduce programming model.

This study was carried out to create a dataset in the project "Cloud Based, Distributed Reputation Analysis Systems for Social Networks", code 2140281, which is supported by Tübitak 1512 Entrepreneurship Progressive Support Program. In the next study, sentiment analysis studies will be done using this dataset.